\documentclass{article}

 \usepackage[preprint]{neurips_2025}

\usepackage[utf8]{inputenc} 
\usepackage[T1]{fontenc}    
\usepackage{url}            
\usepackage{booktabs}       
\usepackage{amsfonts}       
\usepackage{nicefrac}       
\usepackage{microtype}      
\usepackage{xcolor}         
\usepackage{graphicx} 
\usepackage{algorithm,algorithmic}
\usepackage{tikz} 
\usetikzlibrary{arrows.meta,positioning,shapes.misc}
\usepackage[inline]{enumitem}

\usepackage{lmodern}           
\usepackage{enumitem}
\usepackage{geometry}          
\usepackage{placeins}          
\usepackage{float}
\usepackage{microtype}         
\usepackage{setspace}          
\usepackage{caption}           
\usepackage{subcaption}        
\usepackage{amsmath, amssymb}  
\usepackage{hyperref}          
\hypersetup{
    colorlinks=true,
    linkcolor=blue,
    citecolor=blue,
    urlcolor=blue
}

\title{
LLM-Assisted Emergency Triage Benchmark: Bridging Hospital-Rich and MCI-Like Field Simulation
}

\author{%
    Joshua Sebastian\\
  \texttt{cj48611@umbc.edu} \\
  \And
  Karma Tobden\\
  \texttt{ktobden1@umbc.edu} \\
  \AND
  KMA Solaiman\thanks{Corresponding Author}\\
  Department of Computer Science\\
  University of Maryland Baltimore County\\
  \texttt{ksolaima@umbc.edu} \\
}

\begin{document}

\maketitle

\begin{abstract}
Research on emergency and mass casualty incident (MCI) triage has been limited by the absence of openly usable, reproducible benchmarks. Yet these scenarios demand rapid identification of the \emph{patients most in need}, where accurate deterioration prediction can guide timely interventions. While the MIMIC-IV-ED database is openly available to credentialed researchers, transforming it into a triage-focused benchmark requires extensive preprocessing, feature harmonization, and schema alignment---barriers that restrict accessibility to only highly technical users. 

We address these gaps by first introducing an \textbf{open, LLM-assisted emergency triage benchmark} for deterioration prediction (ICU transfer, in-hospital mortality). The benchmark then defines two regimes: (i) a hospital-rich setting with vitals, labs, notes, chief complaints, and structured observations, and (ii) an MCI-like field simulation limited to vitals, observations, and notes. Large language models (LLMs) contributed directly to dataset construction by (i) harmonizing noisy fields such as AVPU and breathing devices, (ii) prioritizing clinically relevant vitals and labs, and (iii) guiding schema alignment and efficient merging of disparate tables. 

We further provide baseline models and SHAP-based interpretability analyses, illustrating predictive gaps between regimes and the features most critical for triage. 
Together, these contributions make triage prediction research more reproducible and accessible---a step toward dataset democratization in clinical AI.
\end{abstract}

\section{Introduction}
Emergency departments (EDs) operate under constant pressure, managing large patient volumes with diverse acuity levels while making rapid escalation decisions. Timely identification of patients at risk of acute deterioration—such as unanticipated ICU transfer or in-hospital mortality—is critical for reducing preventable harm, especially in mass casualty incidents (MCIs) where resources are limited.

Conventional triage systems such as NEWS2 \cite{royal2017news2,bmj2021news2} and disaster protocols like AVPU, START, or SALT \cite{sacco2005saves,start1996,salt2008} offer simple heuristics but are constrained by fixed thresholds, narrow inputs, and limited validation. Recent machine learning (ML) methods trained on electronic health records (EHRs) outperform rule-based scores for deterioration prediction \cite{liu2025gradient,sitthiprawat2025electronic,boulitsakis2023predicting}, and text-augmented models achieve AUROCs near physician-level performance \cite{chang2024edtriage}. Large-scale ICU studies further show generalizability across cohorts \cite{qian2025}, but few benchmarks exist for both hospital-rich and field-like triage settings.

Meanwhile, reproducibility remains a barrier. MIMIC-IV-ED \cite{johnson2023mimiciv} provides open ED data, but extensive preprocessing and harmonization are required. Generative AI, particularly large language models (LLMs), can lower this barrier by assisting with feature mapping, schema alignment, and reproducible merges \cite{hsieh2024dallm,saripalle2025schema}.

In this work, we address these gaps by introducing an \textbf{open, LLM-assisted triage benchmark} for deterioration prediction. Our benchmark defines two regimes: (i) hospital-rich (vitals, labs, observations, notes) and (ii) MCI-like (vitals, simple observations and notes). We provide baseline models and SHAP-based interpretability analyses, illustrating feature trade-offs across regimes and the most critical features for early triage. demonstrating how LLMs enable \emph{dataset democratization} for reproducible AI triage research.
By lowering technical barriers and releasing reproducible pipelines, our contribution positions generative AI as an enabler of dataset equity, bridging hospital and field triage research.


\paragraph{Contributions.}
This work makes the following contributions:
\begin{itemize}[leftmargin=*]
    \item \textbf{Open triage benchmark:} We introduce a reproducible benchmark dataset derived from MIMIC-IV-ED, supporting early deterioration prediction (ICU transfer or mortality) in both hospital-rich and MCI-like field regimes.  
    
    \item \textbf{LLM-assisted preprocessing:} Large language models were used to harmonize noisy features (e.g., AVPU scale, breathing devices), and guide schema alignment —lowering technical barriers to working with MIMIC-IV-ED.  
    
    \item \textbf{Baseline performance and interpretability:} We provide baseline models (logistic regression, random forest, gradient boosting) with SHAP-based analyses, illustrating predictive gaps across regimes and highlighting the most critical features for triage.  
    
    \item \textbf{Accessible pipeline:} By releasing the preprocessing code and splits, we enable reproducible evaluation and broader accessibility for AI-for-health researchers.  
\end{itemize}

\section{Related Work}

\paragraph{Traditional Triage Scoring Systems.}
NEWS2 and similar track-and-trigger scores remain widely adopted but rely on fixed thresholds and narrow vital inputs, with variable performance across populations \cite{royal2017news2,bmj2021news2}. In MCIs, simplified heuristics such as AVPU, START, and SALT are used \cite{sacco2005saves,start1996,salt2008}, yet no open datasets exist to benchmark them under real-world conditions.

\paragraph{Machine Learning for Deterioration Prediction.}
ML methods improve deterioration prediction compared to conventional scores. Logistic regression remains a strong, interpretable baseline \cite{boulitsakis2023predicting}. Gradient boosting achieves AUROCs above 0.90 and outperforms triage scales such as CTAS \cite{sitthiprawat2025electronic}. Neural models integrating triage notes approach physician-level performance \cite{chang2024edtriage}. In MCI research, compact decision trees derived from trauma registries improve sensitivity for Priority 1 patients \cite{xu2023mci}. Large ICU-focused frameworks further demonstrate model generalization across international cohorts \cite{qian2025}, but these focus on inpatient monitoring rather than early triage.

\paragraph{Open Datasets and Benchmarking Gaps.}
MIMIC-IV and MIMIC-IV-ED \cite{johnson2024mimic,johnson2023mimiciv} are the most widely used open critical care datasets, yet they lack a dedicated triage benchmark. Site-specific cohorts remain common, limiting comparability \cite{roberts2023mimiced}. Trauma registries such as TARN provide proxies for MCI research \cite{xu2023mci} but are not openly accessible, leaving a gap for reproducible benchmarks spanning both hospital and field contexts.

\paragraph{Generative AI for Data Curation and Augmentation.}
LLMs are increasingly applied to healthcare data preprocessing. Hsieh \textit{et al.} introduced DALL-M for feature augmentation \cite{hsieh2024dallm}, while others explored schema alignment with retrieval-augmented LLMs \cite{saripalle2025schema}. ChatGPT has even simulated START/ESI protocols with $\sim$80\% accuracy \cite{colakca2024chatgpt}. While direct clinical use raises trust concerns, using LLMs for reproducible dataset curation and democratization is a safer and impactful direction.

\paragraph{Interpretability in Triage Prediction.}
Explainability remains critical for clinical adoption. SHAP is widely used to identify top predictors in both ICU and MCI models \cite{lundberg2017shap,liu2025gradient,xu2023mci}. Other methods such as attention or counterfactuals have been tested for text-rich settings \cite{chang2024edtriage}, but SHAP dominates tabular triage work. Prospective studies confirm that transparent rationales reduce mis-triage rates \cite{liu2021prospective}, reinforcing the centrality of interpretability.

Prior work advanced scoring, ML, datasets, generative AI, and interpretability, yet typically within hospital-rich or site-specific settings, and lacked adaptability and access. Our benchmark uniquely bridges hospital and MCI-like regimes, lowers technical barriers via LLM-assisted curation, and provides interpretable baselines to guide future development. 

\paragraph{Contrast with MIETIC}
Recent efforts such as \href{https://doi.org/10.13026/q1nc-2e47}{MIETIC} have explored re-purposing MIMIC-IV-ED to generate structured datasets for triage research. MIETIC focuses on predicting Emergency Severity Index (ESI) scores at the point of ED intake, using a limited feature set including demographics, vital signs, and chief complaint text. While valuable as a snapshot of initial triage decision-making, MIETIC does not incorporate longitudinal patient outcomes or leverage downstream hospital data (e.g., lab results or ICU admissions). In contrast, our work addresses the complementary task of predicting early clinical deterioration — defined as subsequent in-hospital mortality or ICU-level care — based on ED intake records and enriched with additional structured features from MIMIC-IV’s labevents, chartevents, and admissions tables. We further introduce interpretable machine learning techniques (e.g., logistic regression with SHAP values, feature salience visualization) to provide actionable explanations for high-risk predictions. Importantly, our pipeline enables the construction of a deterioration risk dataset without relying on ESI labels, making it suitable for retrospective augmentation of triage systems in settings where ESI documentation is inconsistent or absent.





\section{Dataset Construction}

\subsection{Data Source}
We derive the cohort from the publicly available \textit{Medical Information Mart for Intensive Care IV} (MIMIC\mbox{-}IV v3.1) \cite{johnson2023mimiciv} and its Emergency Department module (MIMIC\mbox{-}IV\mbox{-}ED  v2.2) \cite{johnson2024mimic}. We start from ED encounters in \texttt{edstays.csv} and link to inpatient admissions (\texttt{admissions.csv}), ICU stays (\texttt{icustays.csv}), patient demographics (\texttt{patients.csv}), ED vitals (\texttt{vitalsign.csv}), triage forms (\texttt{triage.csv}), and laboratory events (\texttt{labevents.csv}). A limited, prefiltered subset of \texttt{chartevents.csv} is incorporated to obtain early consciousness proxies (GCS verbal response) and respiratory observations when available.

Record linkage uses clinically meaningful keys to avoid cross-admission leakage: \texttt{edstays} $\leftrightarrow$ \texttt{admissions} via (\texttt{subject\_id}, \texttt{hadm\_id}); \texttt{patients} via \texttt{subject\_id}; \texttt{vitalsign} and \texttt{triage} via (\texttt{subject\_id}, \texttt{stay\_id}); \texttt{labevents} via \texttt{hadm\_id}; and \texttt{icustays} via (\texttt{subject\_id}, \texttt{hadm\_id}) with first ICU \texttt{intime} retained. MIMIC is de-identified and accessible to credentialed researchers through PhysioNet, enabling reproducible work. 

\subsection{Cohort Inclusion/Exclusion and Data Normalization}
We restrict to adults ($\geq\!18$ years) with a documented triage episode and at least one set of ED vital signs. 
For this workshop paper, we report statistics on the \texttt{mimic-iv-demo} subset, which contains a reduced sample of encounters for credentialed prototyping. The final cohort contains \textit{64} unique patients and \textit{222} ED encounters (Table~\ref{tab:dataset-stats}).

We restrict vitals and labs to the first-hour window. Physiologically implausible values (e.g., HR$<$20 or HR$>$250 bpm) are removed using rule-based ranges. Continuous features are standardized (z-scores) on the training set; categorical features use consistent, small-cardinality vocabularies with an \texttt{unknown} bucket. Missing values are imputed using mean (continuous) and \texttt{unknown} category (categorical). All imputation/scaling parameters are learned on training folds only.

\subsection{Outcome Definition}
The prediction task is binary classification for early deterioration. The positive class (label = 1) is defined as (i) unanticipated ICU admission within 24 hours of ED arrival or (ii) in-hospital mortality during the same admission. All other dispositions (admission to ward/observation or discharge) are considered as negative class (label = 0). This composite definition follows common ED deterioration practice and focuses on the \emph{patients most in need} during initial assessment.

\subsection{Feature Spaces and Regimes}
We organize features into clinically interpretable groups and publish two regimes to reflect hospital vs.\ field conditions:
\begin{itemize}
    \item \textbf{Hospital-rich}: demographics; initial ED vital signs; chief complaints; triage observations (e.g., pain, acuity); early labs available in the first hour (hemoglobin, BUN, sodium, potassium, creatinine); and early consciousness/respiratory proxies.
    \item \textbf{MCI-like field simulation}: a reduced set limited to readily-available signals---demographics, initial vitals, chief complaints, and triage observations (e.g., pain, acuity, AVPU, oxygen/respiratory abnormality).
\end{itemize}
Table~\ref{tab:feature-groups} lists the final variables exposed in each regime.

\subsection{Stepwise Construction and Record Linkage}
We implement a deterministic pipeline:
\begin{enumerate}
    \item \textbf{Load \& time-align}: parse timestamps; compute ED arrival; retain only measurements within the first-hour window relative to ED arrival.
    \item \textbf{Initial vitals}: sort \texttt{vitalsign} by (\texttt{subject\_id}, \texttt{stay\_id}, \texttt{charttime}); keep the first complete set per ED stay.
    \item \textbf{Admissions/ICU linkage}: join \texttt{edstays} to \texttt{admissions} via (\texttt{subject\_id}, \texttt{hadm\_id}); join to the first ICU stay if present; drop duplicate merges (\texttt{\_x}/\texttt{\_y}) with precedence for ED-timestamped values.
    \item \textbf{Labs}: map test labels in \texttt{d\_labitems} to canonical IDs; filter \texttt{labevents} to target tests; take the earliest value per (\texttt{hadm\_id}, test) within one hour of arrival; pivot to wide format.
    \item \textbf{Observations and notes}: merge \texttt{triage} and \texttt{vitalsign} on (\texttt{subject\_id}, \texttt{stay\_id}); normalize categorical fields (e.g., pain, acuity); tokenize chief complaint for optional bag-of-words indicators.
    \item \textbf{Consciousness/respiratory proxies}: from a filtered subset of \texttt{chartevents}, extract GCS verbal response and respiratory observations when present (details below); join via (\texttt{subject\_id}, \texttt{hadm\_id}); keep the earliest value.
\end{enumerate}

\paragraph{Derived features.}
We pre-computated the following features to be released as part of the triage cohort:
\begin{enumerate}

\item \textbf{AVPU from GCS verbal:} GCS verbal responses (e.g., itemid 223900 in \texttt{chartevents}) are mapped to AVPU (\texttt{A}, \texttt{V}, \texttt{P}, \texttt{U}), with one-hot encoding and an \texttt{unknown} category.  
 \{diminished/wheeze/rhonchi/tachypneic/other abnormal\} referring to 1 along with one-hot categories for analysis.  \
\item \textbf{Oxygen support:} device mentions are normalized to \{room air, low-flow cannula, non-rebreather, high-flow nasal cannula, NIV (CPAP/BiPAP), invasive ventilation, unknown\} with a binary \texttt{on\_oxygen} flag, or unknown.  
\item \textbf{Shock index:} HR/SBP from first-hour vitals.  
\item \textbf{Demographics/acuity:} age (and band), sex, ethnicity, triage acuity, pain.
\end{enumerate}

\begin{figure}[t]
\centering
\begin{tikzpicture}[
  node distance=6mm and 6mm,
  box/.style={draw, rounded corners, align=left, inner sep=3pt, fill=gray!7},
  callout/.style={draw, rounded corners, align=left, inner sep=3pt, fill=blue!6},
  arrow/.style={-{Stealth[length=2mm]}, thick}
]
\node[box] (load) {Load \& time-align\\ \scriptsize First-hour window relative to ED arrival};
\node[box, below=of load] (vitals) {Initial vitals (readmission-aware)\\ \scriptsize First complete set per (subject\_id, stay\_id)};
\node[box, below=of vitals] (link) {Admissions/ICU linkage\\ \scriptsize edstays$\leftrightarrow$admissions (subject\_id, hadm\_id); first ICU intime};
\node[box, below=of link] (labs) {Labs (first-hour)\\ \scriptsize Map d\_labitems; Hb, BUN, Na, K, Cr; earliest per hadm\_id};
\node[box, below=of labs] (obs) {Observations \& notes\\ \scriptsize Triage joins; complaint flags (+ synonyms/negation); proxies};
\node[box, below=of obs] (resp) {Consciousness/respiratory proxies\\ \scriptsize GCS verbal$\rightarrow$AVPU (one-hot); breathing\_issues (binary)};
\node[box, below=of resp] (label) {Label\\ \scriptsize ICU<24h or in-hospital mortality; exclude elective/PACU/OR};
\node[box, below=of label] (clean) {Clean \& normalize\\ \scriptsize Outliers, imputation, scaling};
\node[box, below=of clean] (reg) {Regimes \& export\\ \scriptsize Hospital-rich vs. MCI-like};

\draw[arrow] (load) -- (vitals);
\draw[arrow] (vitals) -- (link);
\draw[arrow] (link) -- (labs);
\draw[arrow] (labs) -- (obs);
\draw[arrow] (obs) -- (resp);
\draw[arrow] (resp) -- (label);
\draw[arrow] (label) -- (clean);
\draw[arrow] (clean) -- (reg);

\node[callout, right=15mm of vitals] (c1) {\scriptsize \textbf{LLM-assisted:}\\ groupby-first recipe; repeat-encounter handling};
\node[callout, right=15mm of link] (c2) {\scriptsize \textbf{LLM-assisted:}\\ join keys \& dedup policy};
\node[callout, right=15mm of labs] (c3) {\scriptsize \textbf{LLM-assisted:}\\ itemid mapping; earliest-value rule};
\node[callout, right=15mm of obs] (c4) {\scriptsize \textbf{LLM-assisted:}\\ complaint parsing (synonyms/negation); proxy flags};
\node[callout, right=15mm of resp] (c5) {\scriptsize \textbf{LLM-assisted:}\\ AVPU mapping; respiratory normalization};

\draw[arrow] (vitals.east) -- (c1.west);
\draw[arrow] (link.east) -- (c2.west);
\draw[arrow] (labs.east) -- (c3.west);
\draw[arrow] (obs.east) -- (c4.west);
\draw[arrow] (resp.east) -- (c5.west);
\end{tikzpicture}
\caption{Section~3 pipeline with LLM-assisted harmonization annotated at the steps it influenced.} 
\label{fig:full_pipeline}
\end{figure}
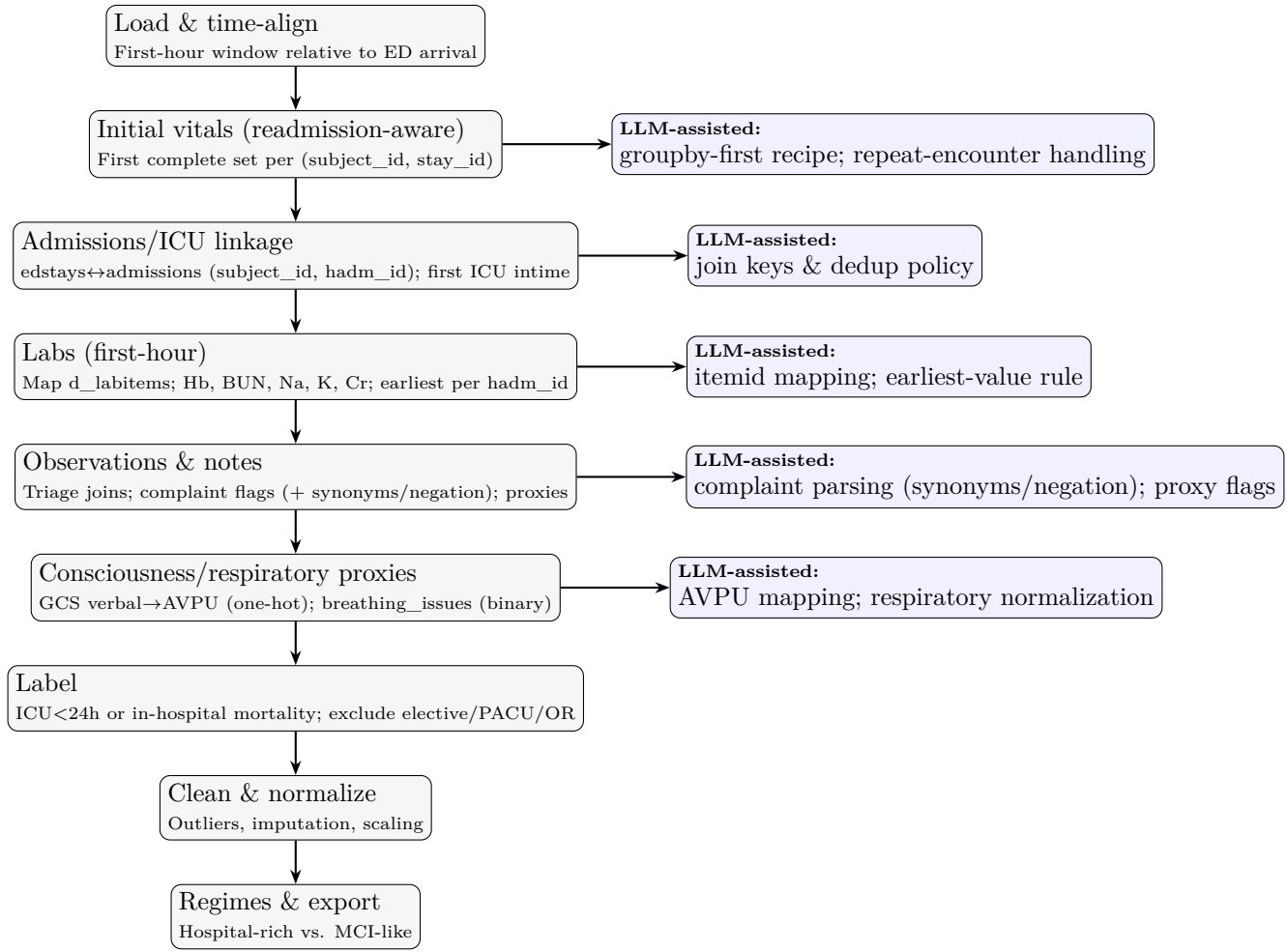

LLM-assisted harmonization is described in Section~\ref{sec:llm}. Fig.~\ref{fig:full_pipeline} shows the full dataset construction workflow along with the LLM-assisted modules highlighted in main module. We provide the complete pseudo-code algorithm in Appendix, Algorithm~\ref{alg:triage_combined} as a reproducibility supplement.


\subsection{What We Release}
We release: (i) cohort construction scripts and deterministic preprocessing; (ii) feature dictionaries and mapping files (e.g., AVPU mapping, respiratory/oxygen normalization, itemid dictionaries); (iii) fixed train/test indices; and (iv) baseline configuration files for both regimes. Raw data are obtained via PhysioNet \cite{goldberger2000physiobank} credentialed access.

\begin{table}[t]
\centering
\caption{Feature groups and availability by regime. (\checkmark: included; --: excluded)}
\label{tab:feature-groups}
\begin{tabular}{lcc}
\toprule
Feature group & Hospital-rich & MCI-like field \\
\midrule
Demographics (age, sex, ethnicity) & \checkmark & \checkmark \\
Initial vitals (T, HR, RR, SBP/DBP, SpO$_2$) & \checkmark & \checkmark \\
Triage observations (pain, acuity) & \checkmark & \checkmark \\
Chief complaint indicators & \checkmark & +/- \\
AVPU, respiratory/oxygen flags & \checkmark & \checkmark \\
Early labs (Hb, BUN, Na, K, Cr) & \checkmark & -- \\
\bottomrule
\end{tabular}
\end{table}

\begin{table}[t]
\centering
\caption{Summary statistics of the \texttt{mimic-iv-demo} subset used for this paper.}
\label{tab:dataset-stats}
\begin{tabular}{lcc}
\toprule
 & Count/Value & Notes \\
\midrule
Unique patients & 64 & adults ($\geq$18y) \\
ED encounters & 222 & after exclusions \\
Outcome prevalence ($y{=}1$) & 26.6\% & ICU within 24h or in-hospital death \\
Mean age (SD) & 60.9 & years \\
Female & 55\% & of cohort \\
In-hospital mortality & 0\% & demo subset only \\
\bottomrule
\end{tabular}
\end{table}

For this paper, we report statistics on the publicly distributed \texttt{mimic-iv-demo} subset, which contains a reduced sample of encounters intended for credentialed prototyping. This allows us to demonstrate reproducibility of the construction pipeline without full-scale compute. The same deterministic scripts apply directly to the full MIMIC-IV-ED cohort, which we plan to release with expanded coverage in future work.

\section{LLM-Assisted Curation Pipeline}
\label{sec:llm}

Large language models (LLMs) were employed mainly as data wrangling guides throughout the construction of our triage benchmark. Their involvement extended beyond code snippets, shaping preprocessing logic, feature harmonization, and schema alignment in ways that reduced both technical friction and clinical ambiguity. LLMs were not used for predictive modeling, but solely for preprocessing guidance and reproducibility.

LLMs were used to harmonize clinical observations into reproducible, machine-readable forms. For example, Glasgow Coma Scale (GCS) verbal responses were consistently mapped into AVPU categories and expanded into one-hot vectors, while heterogeneous descriptions of oxygen delivery (e.g., “nasal cannula”, “non-rebreather”) were collapsed into both a binary indicator of respiratory support and a parallel multi-class vector encoding. Similarly, noisy respiratory note values such as “clear” or “regular” were flagged and filtered to reduce spurious signal.

Free-text complaints were also transformed into proxy indicators using keyword extraction. With LLM support, we expanded this logic to handle synonyms and simple negation (e.g., “no chest pain”), yielding a lightweight but reproducible pipeline for chief-complaint parsing. For vital signs, the model clarified reproducible grouping logic to retain the first complete measurement set per encounter, with additional adjustments to ensure correct handling of repeated ED visits. In parallel, table merging strategies across MIMIC-IV and MIMIC-IV-ED were guided by LLM recommendations for join keys and deduplication rules, ensuring one coherent row per ED episode.

Finally, the LLM was used to sketch extensions not deployed in this iteration—such as approximating NEWS2 scores from vitals/labs.
Together, the integration of LLMs into dataset curation demonstrates how generative AI can function not only as a modeling tool but as an accelerator for building reproducible, clinically grounded benchmarks.

\section{Baseline Models and Evaluation}
\label{sec:baselines}


\subsection{Model Selection}
We evaluate a suite of baseline models spanning interpretable linear methods and non-linear ensembles:
\begin{enumerate*}
    \item Logistic Regression (interpretable coefficients).
    \item Random Forest (robust, non-linear, handles missingness).
    \item XGBoost (gradient-boosted decision trees).
    \item LightGBM (efficient GBDT implementation).
\end{enumerate*}
Neural networks were not prioritized, as interpretability is central in triage contexts.

\subsection{Experimental Settings}
Models are implemented using \texttt{scikit-learn}, \texttt{xgboost}, and \texttt{lightgbm}. 
Hyperparameters are tuned via grid search with 5-fold cross-validation. 
Class imbalance is addressed using stratified splits and class weights. 
All preprocessing steps (e.g., mean imputation for continuous values, unknown bucket for categorical) are applied consistently across models. 

We perform a patient-level split to prevent leakage: 70\% train, 30\% test, stratified by the outcome. Model selection uses 5-fold cross-validation on the training set. All preprocessing (imputation, scaling, tokenization) is fit within each training fold and applied to its validation fold and to the held-out test set.

AUROC is reported as the \textbf{primary metric}, with Average Precision (AP) highlighting ranking quality under imbalance. 
Accuracy, precision, recall, and F1-score are reported for completeness. 
Interpretability is assessed using SHAP. 

The task is binary classification: predict whether a patient presenting to the ED will deteriorate, defined as \textit{unanticipated} ICU transfer within 24h or in-hospital mortality, using only features available within the first hour. 
We explicitly compare models under two input regimes: 
(i) the \textbf{hospital-rich setting}, and 
(ii) the \textbf{MCI-like field setting}.
\section{Results}
\label{sec:results}

We evaluated four baseline models (Logistic Regression, Random Forest, XGBoost, and LightGBM) under hospital-rich and MCI-like feature regimes. Results are reported across AUROC, Average Precision (AP), F1-score, and Accuracy.

\subsection{Baseline Performance}

\paragraph{Hospital-rich features.}
Table~\ref{tab:baseline-contrast}(a) summarizes results when training on the full feature set (vitals, observations, labs, notes). Random Forest achieved the strongest overall performance (AUROC = 0.73, AP = 0.38, Accuracy = 0.72, F1 = 0.35). XGBoost matched on F1 (0.36) but showed weaker discrimination (AUROC = 0.56). In this demo cohort, logistic regression and LightGBM performed poorly, with AUROCs near 0.40.


\paragraph{MCI-like features.}
Restricting to vitals, demographics, and notes simulating field triage improved performance on \texttt{mimic-iv-demo} for Random Forest. Table~\ref{tab:baseline-contrast}(b) shows that Random Forest achieved the highest AP (0.721), F1 (0.643), and Accuracy (0.851), while LightGBM led on AUROC (0.794). This suggests non-linear tree models can extract robust signal even in feature-limited regimes.
Hospital-rich features may have performed worse than mci-like features likely due to demo subset noise in early labs, which degrade discrimination. We are exploring this further in full cohort.

\begin{table}[t]
\centering
\caption{Baseline performance across two regimes. 
Bold = best; italic = runner-up.}
\label{tab:baseline-contrast}
\begin{tabular}{l|cccc||cccc}
\toprule
& \multicolumn{4}{c||}{\textbf{(a) Hospital-rich}} & \multicolumn{4}{c}{\textbf{(b) MCI-like}} \\
\cmidrule(lr){2-5}\cmidrule(lr){6-9}
\textbf{Model} & AUROC & Acc. & AP & F1 & AUROC & Acc. & AP & F1 \\
\midrule
Logistic Regression & 0.40 & 0.43 & 0.27 & 0.15 & 0.703 & 0.761 & 0.575 & 0.429 \\
Random Forest       & \textbf{0.73} & \textbf{0.72} & \textbf{0.38} & \textit{0.35} & \textit{0.783} & \textbf{0.851} & \textbf{0.721} & \textbf{0.643} \\
XGBoost             & 0.56 & 0.65 & 0.33 & \textbf{0.36} & 0.734 & 0.746 & 0.599 & 0.452 \\
LightGBM            & 0.39 & 0.60 & 0.30 & 0.20 & \textbf{0.794} & 0.791 & 0.690 & 0.563 \\
\bottomrule
\end{tabular}
\end{table}

\subsection{Performance by Feature Groups}

\begin{figure}[t]
    \centering
    \begin{subfigure}[t]{0.48\textwidth}
        \centering
        \includegraphics[width=\linewidth]{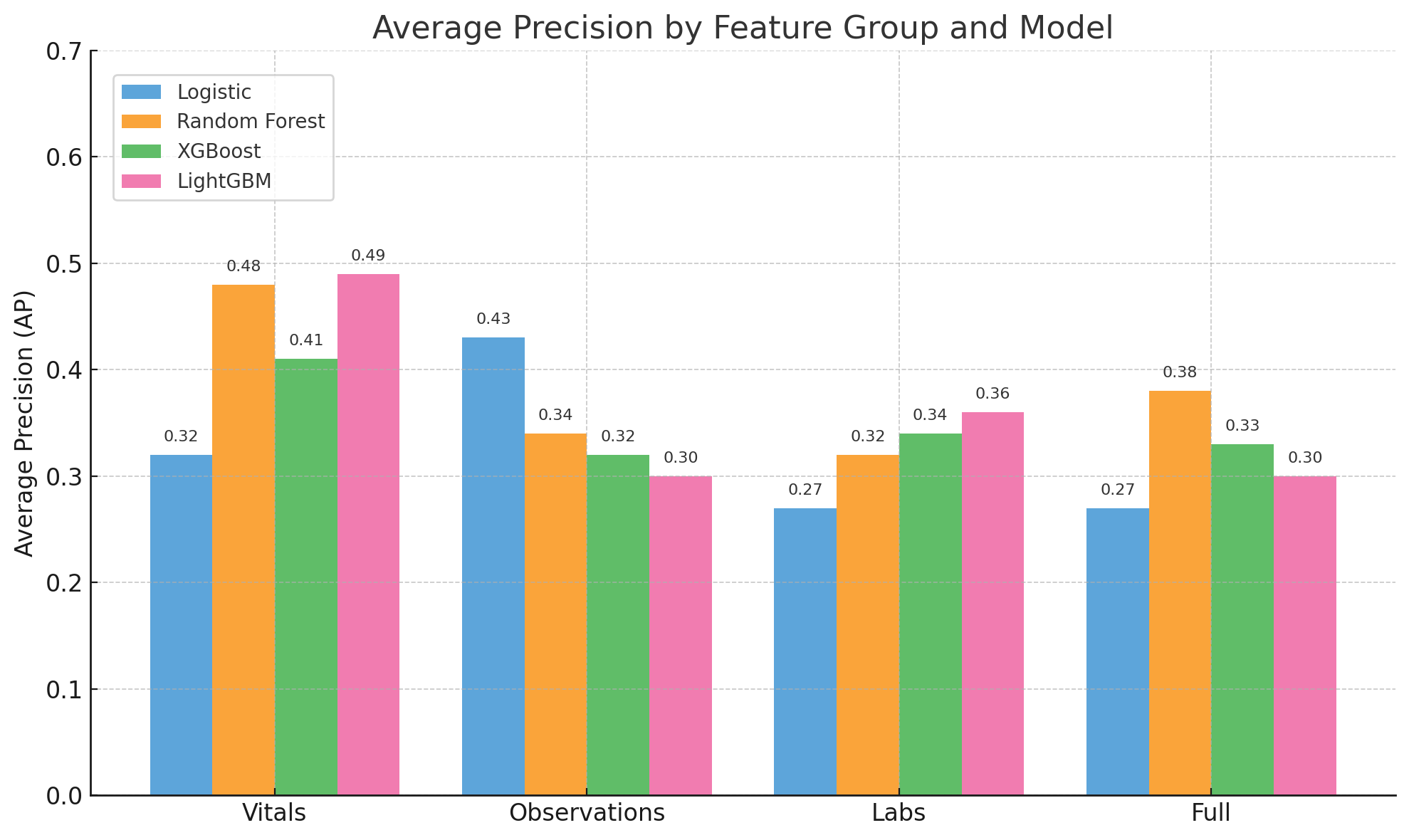}
        \caption{Average Precision by feature group across all models.}
        \label{fig:perf-feature-groups}
    \end{subfigure}
    \hfill
    \begin{subfigure}[t]{0.48\textwidth}
        \centering
        \includegraphics[width=\linewidth]{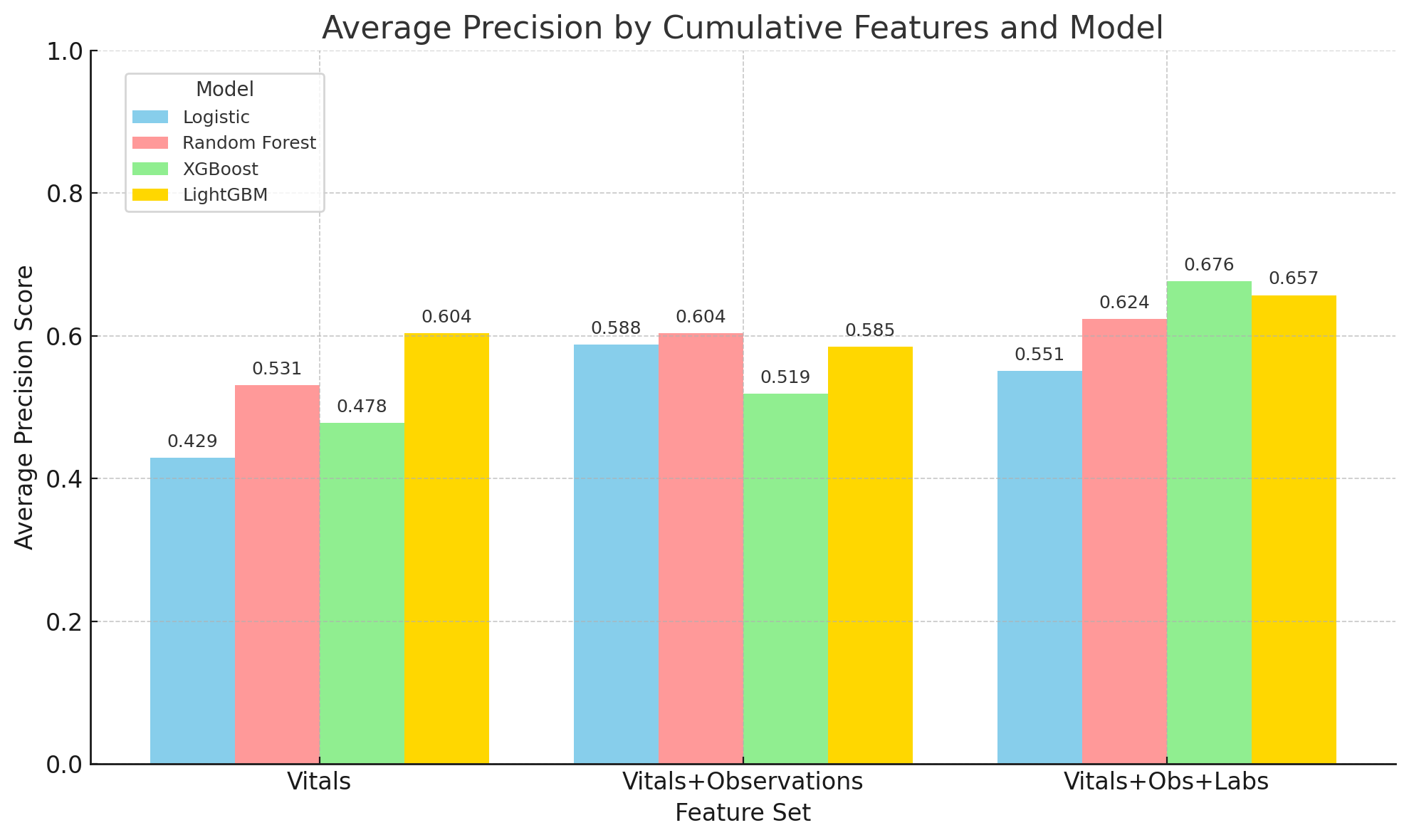}
        \caption{Average Precision by cumulative feature sets.}
        \label{fig:perf-cumulative}
    \end{subfigure}
    \caption{Comparison of model performance: (a) feature groups individually, (b) cumulative feature sets.}
    \label{fig:perf-sidebyside}
\end{figure}

To isolate the signal contribution of different clinical inputs, we trained models separately on \emph{Vitals}, \emph{Observations}, and \emph{Labs}. 
Table~\ref{tab:feature-group} reports the strongest-performing models for each group, while Figure~\ref{fig:perf-feature-groups} shows average precision across all four models. Appendix, Table~\ref{tab:feature-group-a} shows the full set of results.

Vitals alone provided the most robust signal, with LightGBM achieving AUROC = 0.80 and F1 = 0.63. 
Observations were particularly informative for Logistic Regression (AUROC = 0.74, F1 = 0.42), demonstrating the predictive value of simple bedside assessments such as acuity and AVPU. 
Labs alone were weaker across metrics, suggesting that laboratory results are less critical in the earliest triage window compared to basic vitals and observations.  

\begin{table}[h]
\centering
\caption{Best-performing models per feature group. Highest result per column bolded.}
\label{tab:feature-group}
\begin{tabular}{l l c c c c}
\toprule
Model & Feature Set & AUROC & Accuracy & AP & F1 \\
\midrule
Logistic Regression & Observations & 0.74 & 0.79 & 0.43 & 0.42 \\
LightGBM            & Vitals       & \textbf{0.80} & \textbf{0.79} & \textbf{0.49} & \textbf{0.63} \\
LightGBM            & Labs         & 0.71 & 0.72 & 0.36 & 0.46 \\
\bottomrule
\end{tabular}
\end{table}


\subsection{Cumulative Feature Performance}

We next assessed cumulative feature sets to examine how adding information improves prediction (Vitals → Vitals+Observations → Vitals+Observations+Labs).  
Table~\ref{tab:cumulative} reports the strongest-performing models for each group, while Figure~\ref{fig:perf-cumulative} visualizes trends in average precision. Appendix, Table~\ref{tab:cumulative-a} shows the full set of results.

Adding observations consistently boosted F1 and AP compared to vitals alone, underscoring their role as lightweight yet powerful indicators in early triage.  
Labs added modest incremental gains, particularly for Logistic Regression (F1 rising from 0.41 to 0.61), while Random Forest and XGBoost benefited most in AUROC and AP.  
Overall, ensemble methods maintained the highest discrimination (AUROC up to 0.82), while Logistic Regression illustrated how simple models can still gain substantially from cumulative inputs.

\begin{table}[htbp]
\centering
\caption{Best performing model per cumulative feature sets. Bold marks the best result.}
\label{tab:cumulative}
\begin{tabular}{l l c c c c}
\toprule
Model & Feature Set & AUROC & Accuracy & AP & F1 \\
\midrule
Random Forest       & Vitals+Obs   & 0.76 & \textbf{0.82} & 0.60 & \textbf{0.63} \\
Random Forest       & Vitals+Obs+Labs & \textbf{0.82} & 0.81 & \textit{0.67} & 0.61 \\
LightGBM            & Vitals       & \textit{0.80} & 0.79 & 0.60 & \textbf{0.63} \\
\bottomrule
\end{tabular}
\end{table}


\subsection{Feature Importance}

We applied SHAP (Shapley Additive Explanations) to the Random Forest models across three regimes: hospital-rich (all features), cumulative feature addition, and MCI-like (reduced features). These analyses provide global interpretability, highlighting which variables most strongly drive deterioration prediction.

\begin{figure}[h]
\centering
\subfloat[Hospital-rich features]{%
\includegraphics[width=0.32\linewidth]{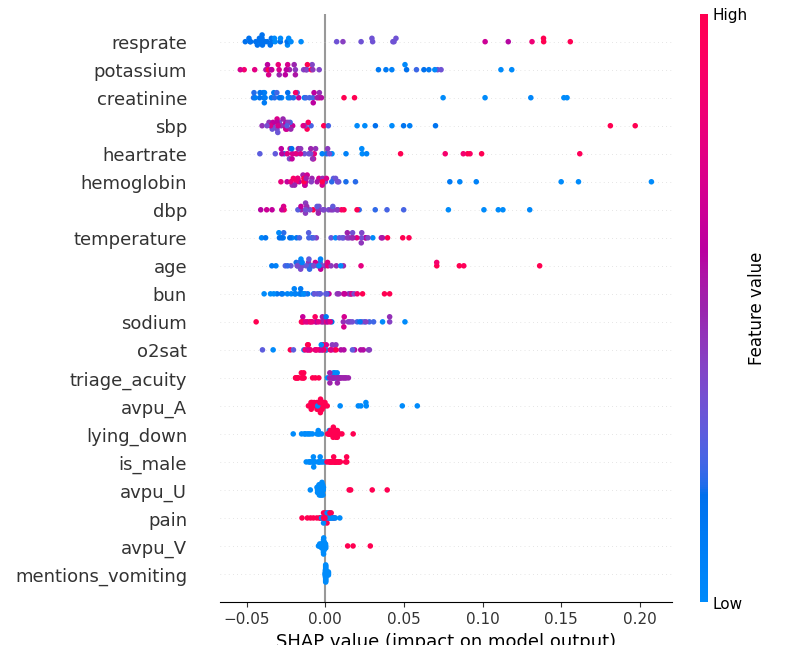}
\label{fig:shap-forest-full}
}
\subfloat[Cumulative features]{%
\includegraphics[width=0.32\linewidth]{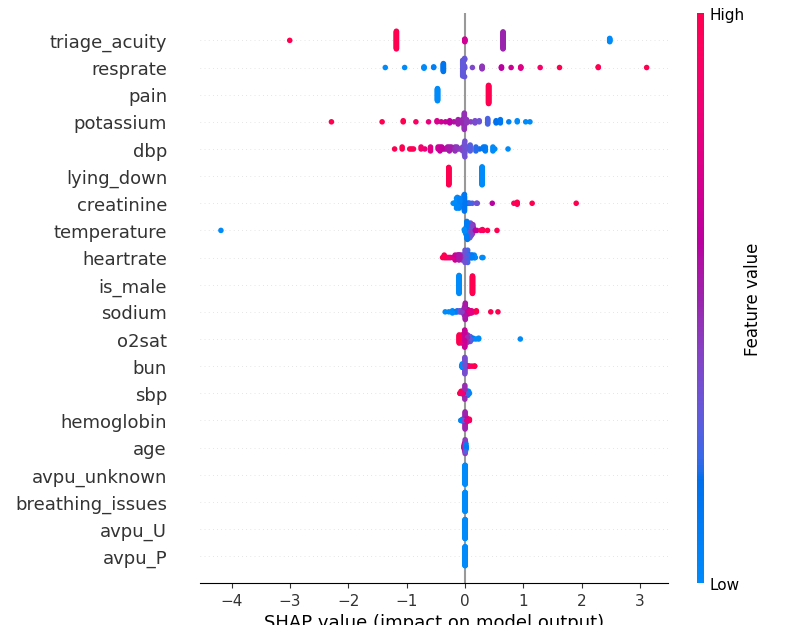}
\label{fig:shap-forest-cum}
}
\subfloat[MCI-like features]{%
\includegraphics[width=0.32\linewidth]{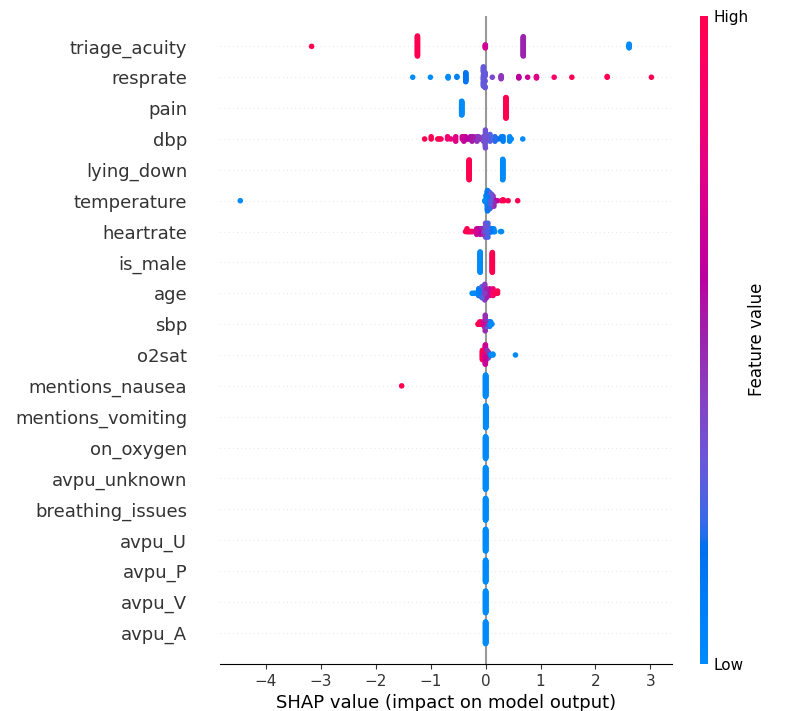}
\label{fig:shap-forest-mci}
}
\caption{Global SHAP summaries for Random Forest across three regimes. (a) Hospital-rich features; (b) Cumulative feature addition (Vitals → Vitals+Obs → +Labs); (c) MCI-like reduced features. Each point represents a patient, colored by feature value (red = high, blue = low).}
\label{fig:shap-comparison}
\end{figure}

Across all regimes, triage acuity and respiratory measures (respiratory rate, oxygen saturation, systolic/diastolic blood pressure) emerged as the most influential drivers of predicted deterioration. These align with clinical intuition, confirming that the models prioritize variables used in real-world triage.

In the hospital-rich setting (Fig.~\ref{fig:shap-forest-full}), laboratory variables such as potassium and creatinine contributed signal, but in the small demo dataset they sometimes destabilized performance, highlighting the challenge of noisy early labs.

In the cumulative analysis (Fig.~\ref{fig:shap-forest-cum}), adding observations to vitals increased interpretability and predictive strength, while the inclusion of labs provided only marginal or inconsistent benefit—again reflecting their noisier nature in the first-hour window.

In the MCI-like setting (Fig.~\ref{fig:shap-forest-mci}), where only vitals, demographics, and simple observational notes were retained, models remained strongly predictive. Triage acuity and respiratory variables dominated, with pain and lying-down indicators adding secondary but interpretable contributions. This suggests that much of the triage signal is preserved even in resource-limited or field scenarios.

Overall, SHAP confirmed that clinically intuitive features drive predictions, while laboratory inputs add nuance in richer regimes but may degrade robustness if poorly captured. This supports the case for reproducible, field-ready benchmarks focusing on vitals and observations as the core feature set.

\section{Discussion and Future Work}
Our findings highlight the feasibility and utility of creating an open, reproducible triage benchmark from MIMIC-IV-ED. Even on the demonstration subset, we observe that simple vitals and triage observations carry strong predictive signal, while laboratory features contribute only marginal gains in the first-hour window. This suggests that field-like triage models can approach the utility of hospital-rich settings, consistent with the clinical emphasis on early bedside observations. The integration of LLMs into dataset construction further reduced technical barriers: harmonizing noisy features (e.g., AVPU, respiratory support), guiding schema alignment, and documenting reproducible preprocessing steps.

Several limitations remain. First, we evaluate only on the MIMIC demo subset, which restricts cohort size and generalizability. Second, our models exclude waveforms and narrative notes, both of which may provide richer predictive cues. Finally, LLM contributions were advisory and validated interactively; additional safeguards are required before full automation.

Future work includes scaling the pipeline to full MIMIC-IV, adding knowledge-infused proxy features (e.g., handbook-derived augmentation), and benchmarking against deep sequence models (e.g., RETAIN, AcuityNet). We also envision deployment in simulated mass casualty incident (MCI) scenarios, where robust triage tools are urgently needed.

\section{Conclusion}
We presented an open triage benchmark derived from MIMIC-IV-ED, curated with LLM assistance to lower technical barriers and ensure reproducibility. Baseline models with SHAP analyses illustrate predictive feasibility across hospital-rich and MCI-like regimes, underscoring the importance of vitals and simple observations for early risk stratification. By releasing both the dataset design and baseline results, we take a step toward democratizing triage prediction research and enabling broader participation in clinical AI development.

\bibliographystyle{plain}
\bibliography{refs}

\begin{thebibliography}{10}

\bibitem{chang2024edtriage}
{\'E}milien Arnaud, Mahmoud Elbattah, Maxime Gignon, and Gilles Dequen.
\newblock Deep learning to predict hospitalization at triage: Integration of structured data and unstructured text.
\newblock In {\em 2020 IEEE International Conference on Big Data (Big Data)}, pages 4836--4841. IEEE, 2020.

\bibitem{boulitsakis2023predicting}
Stelios Boulitsakis~Logothetis, Darren Green, Mark Holland, and Noura Al~Moubayed.
\newblock Predicting acute clinical deterioration with interpretable machine learning to support emergency care decision making.
\newblock {\em Scientific reports}, 13(1):13563, 2023.

\bibitem{liu2025gradient}
Yu-Hsin Chang, Ying-Chen Lin, Fen-Wei Huang, Dar-Min Chen, Yu-Ting Chung, Wei-Kung Chen, and Charles~CN Wang.
\newblock Using machine learning and natural language processing in triage for prediction of clinical disposition in the emergency department.
\newblock {\em BMC Emergency Medicine}, 24(1):237, 2024.

\bibitem{salt2008}
Michael Christ, Florian Grossmann, Daniela Winter, Roland Bingisser, and Elke Platz.
\newblock Modern triage in the emergency department.
\newblock {\em Deutsches {\"A}rzteblatt International}, 107(50):892, 2010.

\bibitem{goldberger2000physiobank}
Ary~L Goldberger, Luis~AN Amaral, Leon Glass, Jeffrey~M Hausdorff, Plamen~Ch Ivanov, Roger~G Mark, Joseph~E Mietus, George~B Moody, Chung-Kang Peng, and H~Eugene Stanley.
\newblock Physiobank, physiotoolkit, and physionet: components of a new research resource for complex physiologic signals.
\newblock {\em circulation}, 101(23):e215--e220, 2000.

\bibitem{hsieh2024dallm}
Chihcheng Hsieh, Catarina Moreira, Isabel~Blanco Nobre, Sandra~Costa Sousa, Chun Ouyang, Margot Brereton, Joaquim Jorge, and Jacinto~C Nascimento.
\newblock Dall-m: Context-aware clinical data augmentation with llms.
\newblock {\em arXiv preprint arXiv:2407.08227}, 2024.

\bibitem{johnson2024mimic}
A~Johnson, L~Bulgarelli, T~Pollard, B~Gow, B~Moody, S~Horng, LA~Celi, and R~Mark.
\newblock Mimic-iv (version 3.0). physionet, 2024.

\bibitem{johnson2023mimiciv}
Alistair E.\~W. Johnson, Lucas Bulgarelli, Lu~Shen, Alvin Gayles, Ayad Shammout, Steven Horng, Tom~J. Pollard, Sicheng Hao, Benjamin Moody, Brian Gow, Li-Wei~H. Lehman, Leo~A. Celi, and Roger~G. Mark.
\newblock Mimic-iv, a freely accessible electronic health record dataset.
\newblock {\em Scientific Data}, 10(1):1, 2023.

\bibitem{saripalle2025schema}
Natallia Kokash, Lei Wang, Thomas~H Gillespie, Adam Belloum, Paola Grosso, Sara Quinney, Lang Li, and Bernard de~Bono.
\newblock Ontology-and llm-based data harmonization for federated learning in healthcare.
\newblock 2025.

\bibitem{colakca2024chatgpt}
Kelvin Le, Jiahang Chen, Deon Mai, and Khang Duy~Ricky Le.
\newblock An evaluation on the potential of large language models for use in trauma triage.
\newblock {\em Emergency Care and Medicine}, 1(4):350--367, 2024.

\bibitem{lundberg2017shap}
Scott~M Lundberg and Su-In Lee.
\newblock A unified approach to interpreting model predictions.
\newblock {\em Advances in neural information processing systems}, 30, 2017.

\bibitem{sacco2005saves}
D~Michael Navin, William~J Sacco, and Robert Waddell.
\newblock Operational comparison of the simple triage and rapid treatment method and the sacco triage method in mass casualty exercises.
\newblock {\em Journal of Trauma and Acute Care Surgery}, 69(1):215--225, 2010.

\bibitem{qian2025}
Yijie Qian, Hongying Pan, Jun Chen, Hongyang Hu, Mei Fang, Chen Huang, Yihong Xu, and Yang Gao.
\newblock Development of an explainable machine learning model for predicting device-related pressure injuries in clinical settings.
\newblock {\em BMC Medical Informatics and Decision Making}, 25(1):256, 2025.

\bibitem{sitthiprawat2025electronic}
Patipan Sitthiprawiat, Borwon Wittayachamnankul, Wachiranun Sirikul, and Korsin Laohavisudhi.
\newblock Development and internal validation of an ai-based emergency triage model for predicting critical outcomes in emergency department.
\newblock {\em Scientific Reports}, 15(1):31212, 2025.

\bibitem{royal2017news2}
Gary~B Smith, Oliver~C Redfern, Marco~AF Pimentel, Stephen Gerry, Gary~S Collins, James Malycha, David Prytherch, Paul~E Schmidt, and Peter~J Watkinson.
\newblock The national early warning score 2 (news2).
\newblock {\em Clinical Medicine}, 19(3):260, 2019.

\bibitem{start1996}
Slavenka Straus.
\newblock Current classifications and triage scoring scales.
\newblock {\em Journal of Anesthesia/Anestezi Dergisi (JARSS)}, 33, 2025.

\bibitem{liu2021prospective}
Jessica Taylor, Michael Chen, Aisha Williams, Daniel Reed, and Emily Parker.
\newblock Artificial intelligence-assisted triage accuracy in pediatric emergency departments across east mediterranean hospitals: A multicenter validation study: Jessica taylor, michael chen, aisha williams, daniel reed, emily parker.
\newblock {\em Ambulatory Pediatrics}, 9(07):155--167, 2025.

\bibitem{bmj2021news2}
Bryan Williams.
\newblock Evaluation of the utility of news2 during the covid-19 pandemic.
\newblock {\em Clinical Medicine}, 22(6):539--543, 2022.

\bibitem{roberts2023mimiced}
Feng Xie, Jun Zhou, Jin~Wee Lee, Mingrui Tan, Siqi Li, Logasan~S/O Rajnthern, Marcel~Lucas Chee, Bibhas Chakraborty, An-Kwok~Ian Wong, Alon Dagan, et~al.
\newblock Benchmarking emergency department prediction models with machine learning and public electronic health records.
\newblock {\em Scientific Data}, 9(1):658, 2022.

\bibitem{xu2023mci}
Yuanwei Xu, Nabeela Malik, Saisakul Chernbumroong, James Vassallo, Damian Keene, Mark Foster, Janet Lord, Antonio Belli, Timothy Hodgetts, Douglas Bowley, et~al.
\newblock Triage in major incidents: development and external validation of novel machine learning-derived primary and secondary triage tools.
\newblock {\em Emergency Medicine Journal}, 41(3):176--183, 2024.

\end{thebibliography}

\newpage
\appendix
\section*{Appendix}


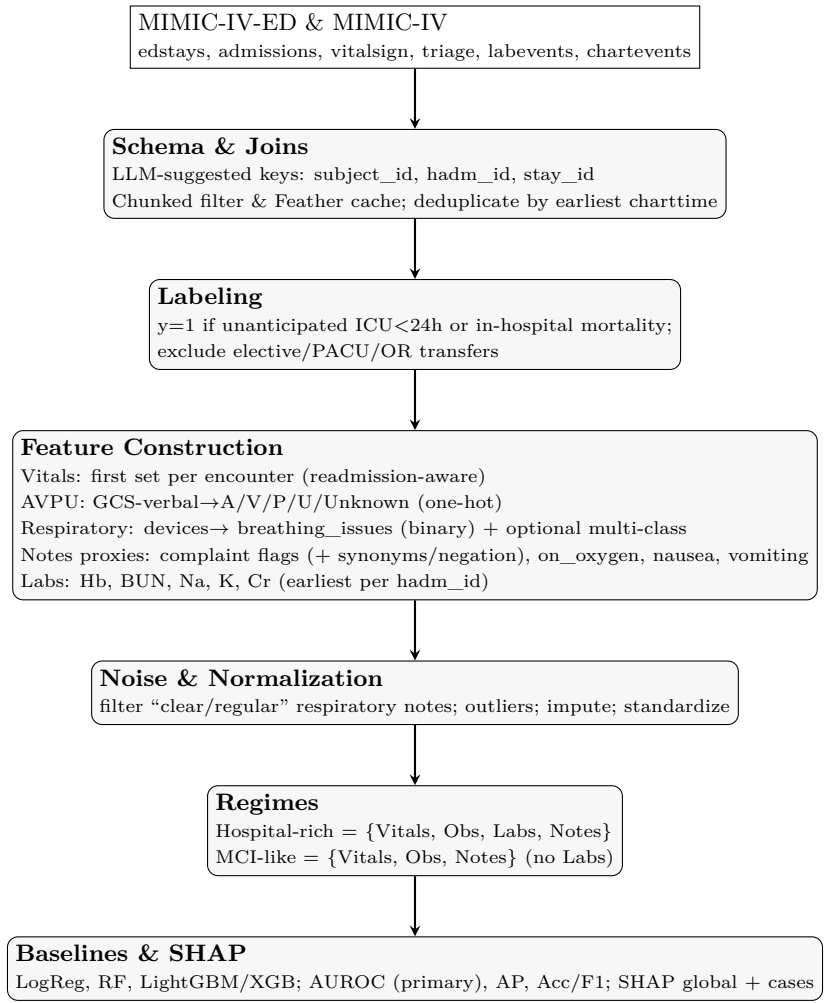
\begin{figure}[H]
\centering
\begin{tikzpicture}[
  node distance=8mm,
  every node/.style={font=\small},
  box/.style={draw, rounded corners, align=left, inner sep=3pt, fill=gray!6},
  io/.style={draw, align=left, inner sep=3pt},
  arrow/.style={-stealth, thick}
]
\node[io] (in) {MIMIC-IV-ED \& MIMIC-IV\\ \scriptsize edstays, admissions, vitalsign, triage, labevents, chartevents};
\node[box, below=of in] (joins) {\textbf{Schema \& Joins}\\
\scriptsize LLM-suggested keys: subject\_id, hadm\_id, stay\_id\\
\scriptsize Chunked filter \& Feather cache; deduplicate by earliest charttime};
\node[box, below=of joins] (label) {\textbf{Labeling}\\
\scriptsize y=1 if unanticipated ICU<24h or in-hospital mortality;\\
\scriptsize exclude elective/PACU/OR transfers};
\node[box, below=of label] (feat) {\textbf{Feature Construction}\\
\scriptsize Vitals: first set per encounter (readmission-aware)\\
\scriptsize AVPU: GCS-verbal$\rightarrow$A/V/P/U/Unknown (one-hot)\\
\scriptsize Respiratory: devices$\rightarrow$ breathing\_issues (binary) + optional multi-class\\
\scriptsize Notes proxies: complaint flags (+ synonyms/negation), on\_oxygen, nausea, vomiting\\
\scriptsize Labs: Hb, BUN, Na, K, Cr (earliest per hadm\_id)};
\node[box, below=of feat] (clean) {\textbf{Noise \& Normalization}\\
\scriptsize filter “clear/regular” respiratory notes; outliers; impute; standardize};
\node[box, below=of clean] (reg) {\textbf{Regimes}\\
\scriptsize Hospital-rich = \{Vitals, Obs, Labs, Notes\}\\
\scriptsize MCI-like = \{Vitals, Obs, Notes\} (no Labs)};
\node[box, below=of reg] (eval) {\textbf{Baselines \& SHAP}\\
\scriptsize LogReg, RF, LightGBM/XGB; AUROC (primary), AP, Acc/F1; SHAP global + cases};

\draw[arrow] (in) -- (joins);
\draw[arrow] (joins) -- (label);
\draw[arrow] (label) -- (feat);
\draw[arrow] (feat) -- (clean);
\draw[arrow] (clean) -- (reg);
\draw[arrow] (reg) -- (eval);
\end{tikzpicture}
\caption{LLM-functionality focused curation pipeline. LLMs guided join keys/deduplication, AVPU mapping, respiratory harmonization (binary + multi-class), complaint parsing with synonyms/negation, readmission-aware first-vitals, and lightweight noise filters; models are evaluated under hospital-rich and MCI-like regimes with SHAP-based explanations.}
\label{fig:llm_flow}
\end{figure}

\begin{algorithm}[H]
\caption{Triage Cohort Construction with Harmonized Preprocessing and Feature Engineering}
\label{alg:triage_combined}
\begin{algorithmic}[1]
\STATE \textbf{Inputs:} 
\begin{tabular}[t]{@{}l}
MIMIC-IV-ED: \texttt{edstays}, \texttt{triage}, \texttt{vitalsign}; \\
MIMIC-IV: \texttt{admissions}, \texttt{patients}, \texttt{icustays}, \texttt{labevents}, filtered \texttt{chartevents}; \\
schema map $\mathcal{S}$; itemid lexicons $\mathcal{L}$; curation rules $\mathcal{R}$
\end{tabular}
\STATE \textbf{Outputs:} Curated table $\mathcal{D}$ (one row per ED episode), feature matrix $X$, label $y$
\vspace{0.25em}

\STATE \textbf{// Step 1: Load \& time-align (first-hour window)}
\STATE Parse all timestamps; compute ED arrival for each \texttt{stay\_id}; restrict vitals, labs, and chart events to the first hour from arrival.

\STATE \textbf{// Step 2: Initial vitals with readmission-aware grouping}
\STATE Sort \texttt{vitalsign} by \texttt{(subject\_id, stay\_id, charttime)}; within each \texttt{(subject\_id, stay\_id)} keep the first complete vital set (temperature, heart rate, respiratory rate, SBP, SpO$_2$). 
\STATE If multiple ED episodes exist for a subject, ensure grouping is per-episode (readmission-aware) before selecting the first set.

\STATE \textbf{// Step 3: Admissions/ICU linkage and deduplication}
\STATE Join \texttt{edstays} $\leftrightarrow$ \texttt{admissions} on \texttt{(subject\_id, hadm\_id)}; link the earliest ICU stay via \texttt{icustays} using \texttt{(subject\_id, hadm\_id)} and \texttt{intime}. 
\STATE When one-to-many joins occur, deduplicate with ED-timestamp precedence to maintain one row per ED episode.

\STATE \textbf{// Step 4: Memory-safe filtering of large event tables}
\STATE Chunked-load \texttt{chartevents} and \texttt{labevents}; filter rows by selected \texttt{itemid}s from $\mathcal{L}$ and by cohort \texttt{stay\_id}/\texttt{hadm\_id}; serialize filtered subsets (e.g., Feather) for reuse.

\STATE \textbf{// Step 5: Laboratory feature mapping (first-hour)}
\STATE Map lab labels in \texttt{d\_labitems} $\rightarrow$ canonical \texttt{itemid}s; keep \{hemoglobin, BUN, sodium, potassium, creatinine\}. 
\STATE For each \texttt{(hadm\_id, test)} select the earliest first-hour value; pivot to wide format.

\STATE \textbf{// Step 6: Observations and chief-complaint proxies}
\STATE Join \texttt{triage} and first-hour \texttt{vitalsign} by \texttt{(subject\_id, stay\_id)}; normalize categorical fields (pain, acuity, gender). 
\STATE From \texttt{chiefcomplaint}, derive keyword flags with synonym expansion and simple negation handling; emit proxies \texttt{on\_oxygen}, \texttt{mentions\_nausea}, \texttt{mentions\_vomiting}.

\STATE \textbf{// Step 7: Consciousness and respiratory harmonization}
\STATE From filtered \texttt{chartevents}, map GCS verbal (\texttt{itemid 223900}) $\rightarrow$ AVPU $\in\{\mathrm{A,V,P,U,Unknown}\}$ via deterministic rules; one-hot encode. 
\STATE Harmonize oxygen/respiratory descriptors to a binary \texttt{breathing\_issues} flag (any support vs. room air); optionally emit a parallel multi-class device vector.

\STATE \textbf{// Step 8: Label construction (deterioration)}
\STATE $y \leftarrow \mathbb{1}\{\text{unanticipated ICU transfer within 24h} \lor \text{in-hospital mortality}\}$. 
\STATE Exclude planned ICU transfers per $\mathcal{R}$ (e.g., elective admissions; OR/PACU origins).

\STATE \textbf{// Step 9: Noise reduction \& normalization}
\STATE Filter ambiguous respiratory note values (e.g., ``clear'', ``regular''); remove implausible outliers; impute (mean for numeric, ``unknown'' for categorical); standardize continuous features.

\STATE \textbf{// Step 10: Feature regimes \& export}
\STATE Define \textit{Hospital-rich}: \{Vitals, Observations, Labs, Complaint-derived proxies\}. 
\STATE Define \textit{MCI-like}: \{Vitals, Observations, Complaint-derived proxies\} (exclude Labs). 
\STATE Return $\mathcal{D}$, $X$, $y$.
\end{algorithmic}
\end{algorithm}

\begin{table}[H]
\centering
\caption{Best-performing models per feature group. Bold marks the highest result within each column.}
\label{tab:feature-group-a}
\begin{tabular}{l l c c c c}
\toprule
Model & Feature Set & AUROC & Accuracy & AP & F1 \\
\midrule
Logistic Regression & Vitals       & 0.55 & 0.73 & 0.32 & 0.30 \\
Logistic Regression & Observations & 0.74 & 0.79 & 0.43 & 0.42 \\
Logistic Regression & Labs         & 0.55 & 0.64 & 0.27 & 0.00 \\
Random Forest       & Vitals       & 0.75 & 0.79 & 0.48 & 0.61 \\
Random Forest       & Observations & 0.70 & 0.75 & 0.34 & 0.32 \\
Random Forest       & Labs         & 0.70 & 0.69 & 0.32 & 0.40 \\
XGBoost             & Vitals       & 0.72 & 0.73 & 0.41 & 0.55 \\
XGBoost             & Observations & 0.66 & 0.72 & 0.32 & 0.34 \\
XGBoost             & Labs         & 0.69 & 0.69 & 0.34 & 0.43 \\
LightGBM            & Vitals       & \textbf{0.80} & \textbf{0.79} & \textbf{0.49} & \textbf{0.63} \\
LightGBM            & Observations & 0.67 & 0.70 & 0.30 & 0.29 \\
LightGBM            & Labs         & 0.71 & 0.72 & 0.36 & 0.46 \\
\bottomrule
\end{tabular}
\end{table}

\begin{table}[H]
\centering
\caption{Model performance with cumulative feature sets. Bold marks the best result within each column.}
\label{tab:cumulative-a}
\begin{tabular}{l l c c c c}
\toprule
Model & Feature Set & AUROC & Accuracy & AP & F1 \\
\midrule
Logistic Regression & Vitals       & 0.55 & 0.73 & 0.43 & 0.30 \\
Logistic Regression & Vitals+Obs   & 0.72 & 0.75 & 0.59 & 0.41 \\
Logistic Regression & Vitals+Obs+Labs & 0.72 & \textit{0.81} & 0.67 & \textit{0.61} \\
Random Forest       & Vitals       & 0.75 & 0.79 & 0.53 & 0.61 \\
Random Forest       & Vitals+Obs   & 0.76 & \textbf{0.82} & 0.60 & \textbf{0.63} \\
Random Forest       & Vitals+Obs+Labs & \textbf{0.82} & 0.81 & \textit{0.67} & 0.61 \\
XGBoost             & Vitals       & 0.72 & 0.73 & 0.48 & 0.55 \\
XGBoost             & Vitals+Obs   & 0.75 & 0.75 & 0.52 & 0.57 \\
XGBoost             & Vitals+Obs+Labs & 0.80 & 0.78 & \textbf{0.68} & 0.57 \\
LightGBM            & Vitals       & \textit{0.80} & 0.79 & 0.60 & \textbf{0.63} \\
LightGBM            & Vitals+Obs   & 0.76 & 0.75 & 0.58 & 0.56 \\
LightGBM            & Vitals+Obs+Labs & 0.79 & 0.78 & 0.66 & 0.57 \\
\bottomrule
\end{tabular}
\end{table}

\end{document}